\documentclass{bmvc2k}

\usepackage{listings}

\title{Fast Dense Feature Extraction with CNNs that have Pooling or Striding Layers}

\addauthor{Christian Bailer}{christian.bailer@dfki.de}{1}
\addauthor{Tewodros A. Habtegebrial}{tewodros_amberbir.habtegebrial@dfki.de}{2}

\addauthor{Kiran Varanasi}{kiran.varanasi@dfki.de}{1}
\addauthor{Didier Stricker\textsuperscript{1,}}{didier.stricker@dfki.de}{2}

\addinstitution{
 German Research Center for Artificial Intelligence (DFKI)\\
 Kaiserslautern, DE
}
\addinstitution{
 University of Kaiserslautern\\
 Kaiserslautern, DE
}

\runninghead{Bailer et. al}{Fast Feature Extraction with CNNs with Pooling Layers}

\begin{document}

\maketitle

\begin{abstract}
In recent years, many publications showed that convolutional neural network based features can have a superior performance 
to engineered features. However, not much effort was taken so far to extract local features efficiently for a whole image.
In this paper, we present an approach to compute patch-based local feature descriptors efficiently in presence of pooling and striding layers for whole images at once. 
Our approach is generic and can be applied to nearly all existing network architectures.
This includes networks for all local feature extraction tasks like camera calibration, Patchmatching, optical flow estimation and stereo matching.
In addition, our approach can be applied to other patch-based approaches like sliding window object detection and recognition. 
We complete our paper with a speed benchmark of popular CNN based feature extraction approaches applied on a whole image, with and without our speedup,
and example code (for Torch) that shows how an arbitrary CNN architecture can be easily converted by our approach.  

\end{abstract}

\begin{figure}
 \includegraphics[width=1.00\linewidth]{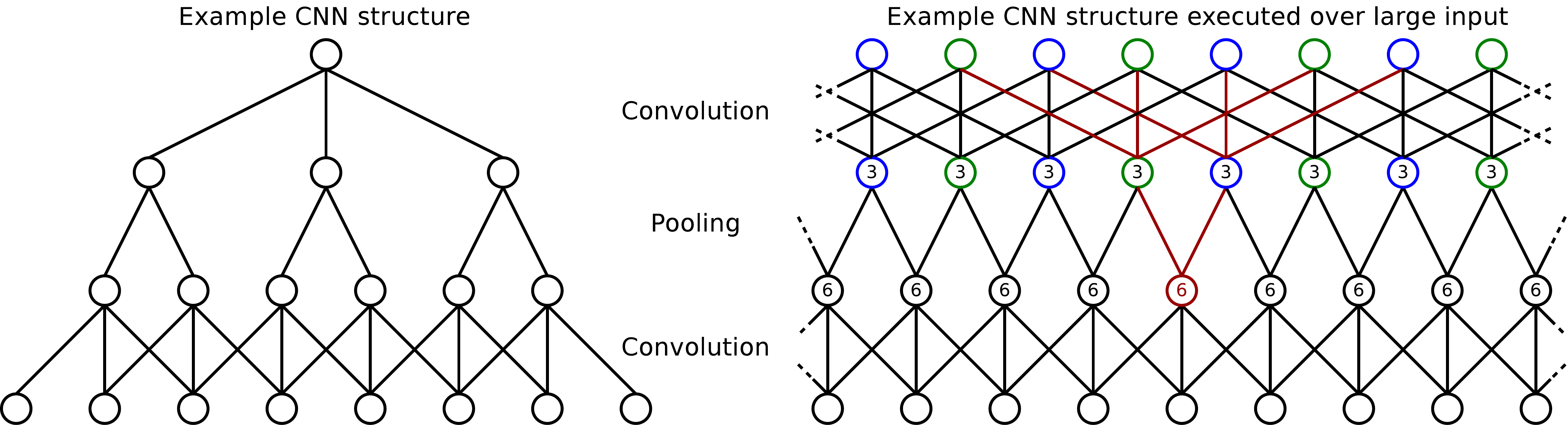}
 \caption{Left: A simple 1 dimensional CNN.
   Right: if this CNN is executed at each pixel position of an image to create features for every position many intermediate layer results are shared between networks.
   The numbers in nodes state how often a node is shared. The red connections show how the red node is shared. 
   Pooling with stride 2 halves the output resolution. Thus, we need two pooling layers: the original one (blue) and one shifted by one pixel (green)
   to avoid halving the output resolution.}
   \label{mainfig}
\end{figure}

\section{Introduction}
\label{sec:intro}
While most CNNs are directly executed on complete images, there are also many important tasks that require 
patch based CNN processing i.e. executing the same CNN several times on neighboring, overlapping patches in an image.
Most of these tasks fall into the category of CNN based feature extraction~\cite{han2015matchnet,simo2015discriminative}.
This includes tasks like camera calibration, Patchmatching~\cite{barnes2010generalized}, optical flow estimation~\cite{bailer2016cnn,gadot2016patchbatch} and stereo matching~\cite{zbontar2016stereo}.
However, there are also important patch based applications that are often not considered as feature extraction tasks like sliding window object detection or recognition~\cite{girshick2015fast}. 

In all such patch based tasks there can be a lot of redundancy between the computations of neighboring CNNs, as shown in Figure~\ref{mainfig}.
If there are no pooling or striding layers this redundancy in calculation can easily be avoided by simply executing 
a CNN which was trained on a limited patch, directly once on the full image. 
However, with pooling layers the situation is more complex.
So far, authors avoided pooling or striding layers completely~\cite{zbontar2016stereo}, 
simply performed the redundant calculations~\cite{gadot2016patchbatch}, 
or designed the approach in a way that it can also work with more sparse results~\cite{girshick2015fast,ge2015modelling}.
The only work that we are aware of, that tries to avoid the redundancy is our previous work~\cite{bailer2016cnn}, 
where the method for avoiding this redundancy was not detailed.

In this paper, we present an elegant and generalizable solution to avoid this redundancy even in the presence of
pooling or striding layers. 
Our approach requires only layers performing transpose and reshape operations, in addition to ordinary CNN layers.
Such operations are available in nearly  all machine learning frameworks.
Furthermore, our approach can be applied on nearly every existing CNN architecture. 

Our paper is structured as follows: after presenting related work in Section~\ref{rel}, we present our approach in Section~\ref{apr}.
A benchmark of our approach is performed in Section~\ref{exp}. Finally, in the appendix we present example source code 
for the deep learning framework Torch to make our contribution even clearer. 

\section{Related Work}\label{rel}
Besides extracting robust feature descriptors there was always also the need to compute these features fast.
A prominent example for this with engineered features is SURF~\cite{bay2006surf}. 
While its predecessor SIFT~\cite{lowe1999object} uses Gaussian filters, SURF uses mean filters. 
This allows very fast dense feature extraction with integral images at the cost of robustness.

In recent years, features based on convolutional neural 
networks~\cite{bailer2016cnn,han2015matchnet, simo2015discriminative, zagoruyko2015learning, simonyan2014learning} showed not only promising, but
mostly even superior results to engineered features.
Zagoruyko and Komodakis~\cite{zagoruyko2015learning} compared  different architectures to compare image patches. 
While they did not perform a speed comparison they noticed that 
the Siamese architecture~\cite{bromley1993signature} with $L_2$ distance is much faster than 2-channel based approaches.
While the 2-channel architecture requires running a matching CNN for every feature comparison, the Siamese architecture
only needs to run a CNN to create a feature but not to match. Once it is created it can be matched by $L_2$ distance.
Simo-Serra et al.~\cite{simo2015discriminative} further exploited this idea.
However, like~\cite{zagoruyko2015learning} they just executed their approach on a predefined set of patches and did not present a fast way 
to compute their features densely on a whole image. We show in this paper that their approach can be adjusted accordingly. 

To perform dense stereo matching Zbontar et. al~\cite{zbontar2016stereo} had to compute features densely on the whole image.
They avoided pooling and striding layers by using CNNs only on tiny patches.  While CNNs on smaller patches are less robust by themselves,
techniques like semi global matching for regularization  still allowed them to get state-of-the art results. 
In contrast, for tasks that cannot be that well regularized like optical flow estimation or image calibration, 
one has to either follow our approach or do the redundant calculations. 

In the application of object recognition the issue of fast dense feature extraction was avoided by simply extracting features sparsely 
and using regression to find the exact object bounding box~\cite{girshick2015fast}. 
While this approach is powerful we think that a real dense approach could still improve results, especially
in regions where many objects heap or simply to process also less interesting regions with a powerful CNN.

\begin{figure}[h]
\centering
 \includegraphics[width=0.55\linewidth]{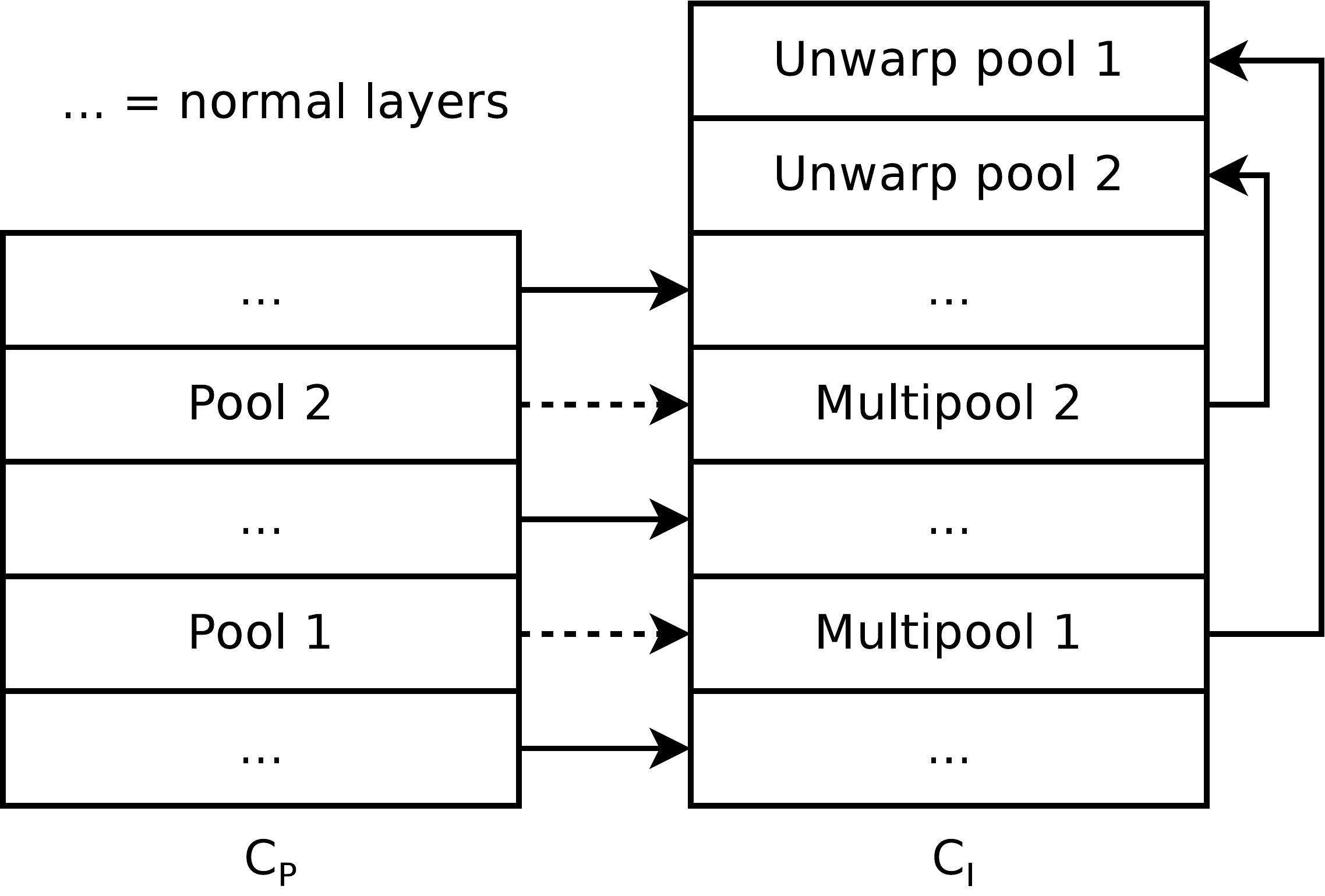}
 \caption{In our approach we create the network $C_I$ from network $C_P$. $C_I$ gives the same result as executing the network $C_P$ on
 every patch of the image $I$, independently.  However, $C_I$  runs much faster as it avoids redundancy between overlapping patches.}
 \label{arch}
\end{figure}

\section{Approach}\label{apr}
In this section we describe our approach.
If we have an input image $I$ with width $I_w$ and height $I_h$, 
we can define patches $P(x,y)$ with width $P_w$ and height $P_h$ centered at each pixel position $(x,y)$, $x\in 0 ... I_w-1,y\in 0 ... I_h-1$
in the input image $I$. Patches lying at the image boundary like $P^I(0,0)$ require image padding as they contain pixels outside
the image area. Still it is common to include such patches to be able to extract features for boundary pixels, as well. 
Of course it is also possible to only consider patches lying 100\% inside the image area. 
However, for simplicity we assume that there is a patch surrounding each image pixel.

In this paper we want to efficiently execute a CNN $C_P$ (which was trained on training 
patches $P^T$) on all patches $P(x,y)$ in the input image $I$ at once. 
The output vector $O(x,y) = C_P(P(x,y))$ 
is a $k$ channel vector which belongs to the $(I_h,I_w,k)$ dimensional output matrix $O$ that 
contains the results of $C_P$ executed on all image patches $P(x,y)$.

Our goal is to create a network $C_I$ that directly calculates $O$ from $I$, while avoiding the redundancy that occurs when 
$C_P$ is executed on each image patch independently. The architectural differences between $C_P$ and $C_I$ are shown in Figure~\ref{arch}.
In the remainder of this sections we describe the steps necessary to get from $C_P$ to $C_I$, namely: in Section~\ref{s21}  we describe 
how to deal with ordinary layers (without pooling or striding). In Section~\ref{s22}  we detail how we create multipooling from pooling and finally in
Section~\ref{s23} we show how mulipooling can be unwarped again. As notation for single layers of $C_I$ and $C_P$ we use $L_I$ and $L_P$, respectively.

\subsection{Layers without pooling}\label{s21}

With no striding or pooling the layers of $C_P$ and $C_I$ are identical i.e. $L^{nopool}_P = L^{nopool}_I$.
This is because their output does not depend on the spacial position of the input, but only on the 
input values itself. 

\begin{figure}[h]
 \includegraphics[width=1.00\linewidth]{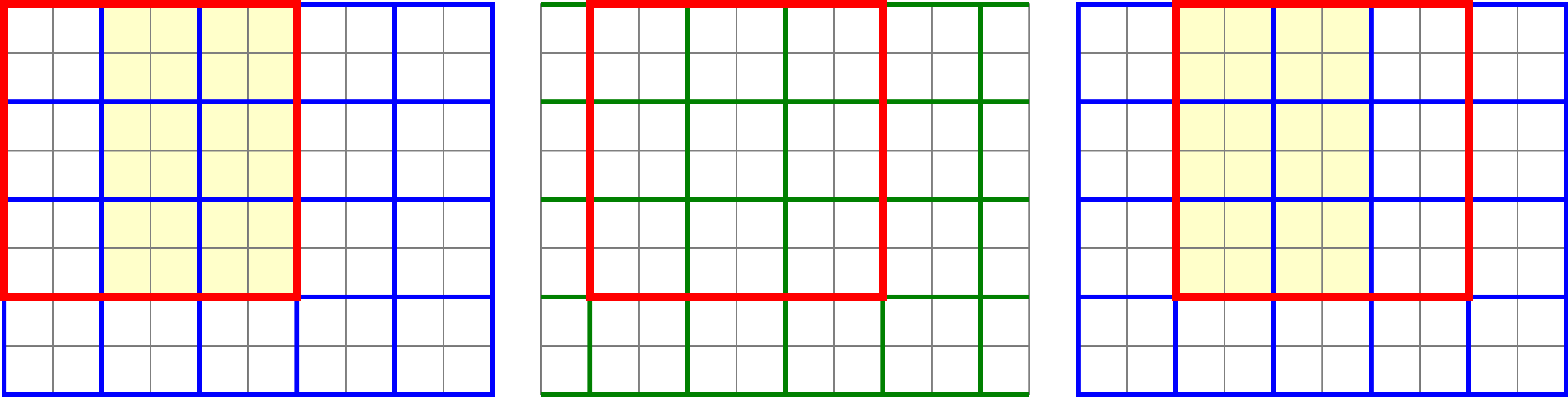}
 \caption{Patches $P$ at different image positions (in red). The first patch $P(x,y)$ requires different 2x2 pooling (blue) than the second patch $P(x+1,y)$ (green). However, the patch $P(x+2,y)$ can work 
 with the original pooling again (blue). Overlapping positions of $P(x,y)$ and $P(x+2,y)$ provide identical results and can thus be shared (bright yellow). Sharing 
 between patches that are using blue and the ones that are using green pooling is not possible. }
 \label{fstride}
\end{figure}

\subsection{Multipool to consider all locations}\label{s22}
In contrast to ordinary layers striding and pooling layers must be handled explicitly. 
However, the kind of pooling has no influence on the handling and striding can be seen as a special kind of pooling layer.
Also, it does not make a difference if the pooling layer is executed directly on the input image $I$ or the outputs of one or several preceding
 layers. If there are preceding layers we simply get a different input $I^*$ with patches $P^*(x,y)$ that can be processed by 
the remaining layers $C^*_P$ or $C^*_I$, receptively.

Figure~\ref{fstride} visualizes the main issue of pooling: different patches $P(x,y)$ require different poolings even if they are direct neighbors like
$P(x,y)$ and $P(x+1,y)$ and can thus not share pooling outputs. 
However, with $s$ being the pooling/stride size and $u$ and $v$ being integers the patches $P(x,y)$ and $P(x+su,y+sv)$ still share pooling outputs
for pixels that are shared by both patches (yellow area in Figure~\ref{fstride}). 
This creates all together $s\times s$ different pooling situations that have to be computed independently on the input $I^L$ of our pooling layer.
As a $s\times s$ pooling layer reduces the output size to $I_w/s,I_h/s$ (with input size $I_w,I_h$) it is clear that $s\times s$ such outputs are required 
to still obtain an output $O$ of spacial size $I_w,I_h$.

With $SHIFT_{y,x}(I)$ being a shifting operation that shifts the input $I$ by $x$ pixels rightwards and $y$ pixels downwards
and $Pool_{s \times s}$ being a pooling operation with stride $s \times s$ we can define a shifted pooling operation:
\begin{equation}
  Pool_{s \times s}^{x,y}(I) =  Pool_{s \times s}(SHIFT_{y,x}(I))
\end{equation}
  A set of shifted pooling operations with $s \times s$ shift distances we call multipooling.
  To convert $C_P$ to $C_I$ we have to replace pooling layers $L^{pool}_P \in C_P$  by multipooling layers $L^{multipool}_I \in C_I$:
\begin{equation}
  L^{multipool}_I =  \{ Pool_{s \times s}^{0,0}, Pool_{s \times s}^{0,1}, ..., Pool_{s \times s}^{0,s-1},..., Pool_{s \times s}^{s-1,0},..., Pool_{s \times s}^{s-1,s-1} \}
\end{equation}
The different pooling outputs are stacked in an extra output dimension 
which we call $M$. Samples in $M$ are treated as independent samples by subsequent layers (similar to a batch dimension).  
Note that $M$ actually consists of two dimensions $M=(y,x)$ as the multipooling contains $y$ as well as $x$ shift 
(and the $y$ shift only increases after processing all $x$ shifts once).
If there is more than one pooling layer a subsequent pooling layer replicates the input dimension $M_{in}$ $s \times s$ times for $M_{out}$ i.e.
$M$ will then consist of $M= (y_n,x_n,...,y_1,x_1)$ after $n$ pooling layers with $y_1,x_1$ belonging to the first pooling layer and $y_n,x_n$ to the $n$th pooling layer.

\begin{figure}[h]
\centering
 \includegraphics[width=0.70\linewidth]{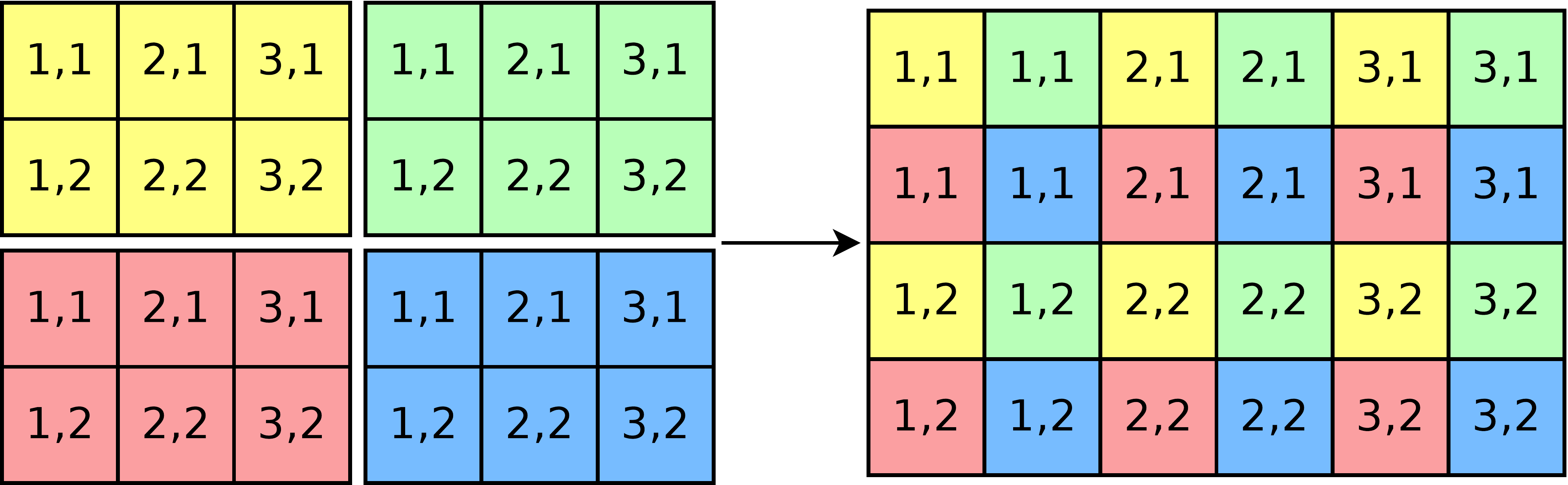}
 \caption{Left: $2 \times 2$ = 4 output images from $2 \times 2$ multipooling. Right: the final unwarping output $O$. We present a generic and efficient way
 of unwarping in Section~\ref{s23}.}
 \label{funwrap}
\end{figure}

\begin{figure}[h]
\centering
 \includegraphics[width=0.80\linewidth]{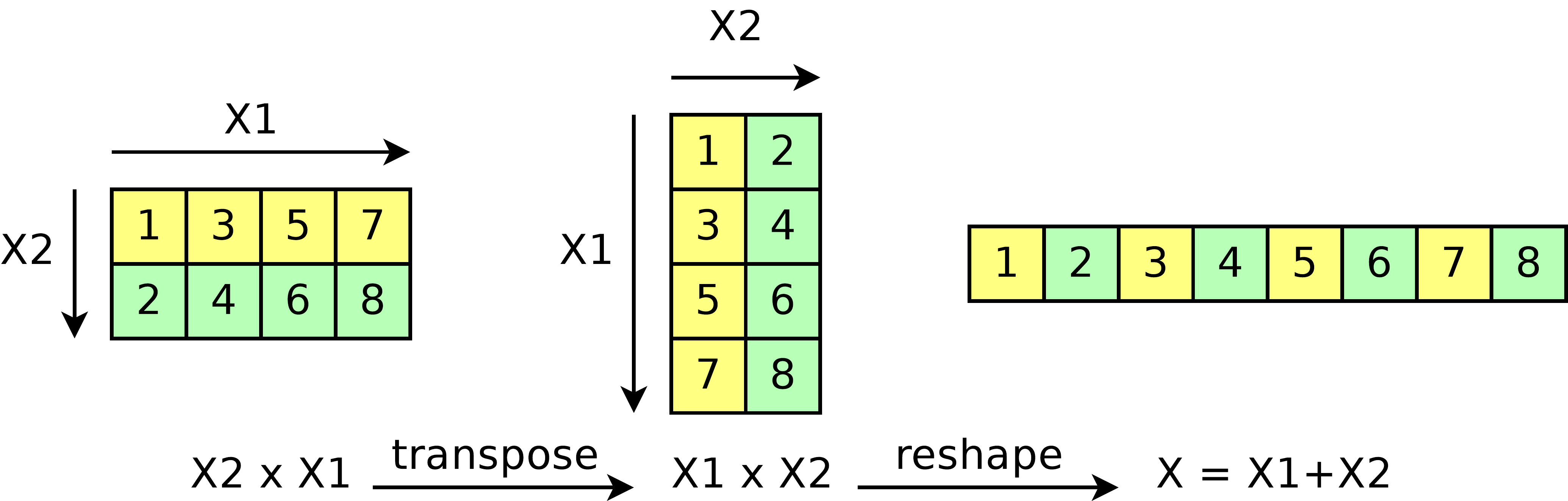}
 \caption{ In the problem $(x_2,x_1+1) = (x_2,x_1)+2$ and $(x_2,x_1) = (x_2,x_1)+1$ i.e. the step size for the inner dimension $x_1$ is larger.
  This can be fixed by transposing (swapping)  both dimensions .
  A reinterpretation of the memory (reshaping) allows then to reduce it to a single $x$ dimension. 
 }
 \label{funwrape}
\end{figure}

\subsection{Unwarping}\label{s23}

With one multipool layer, we get an output $W$ with dimensions $W=(M = s \times s $, $I_h/s, I_w/s, k)$, which  we want to unwarp to the final output $O=(I_h,I_w,k)$.
Figure~\ref{funwrap} shows this unwarping for $2 \times 2$ pooling.
Direct unwarping is complex especially with several pooling layers. This might be a reason why previous work avoided pooling layers.
However, if we observe the problem in dimension space it can easily be solved with solely transpose and reshape operations. Such operations are 
supported by most deep learning frameworks as layers. 

Let us denote   $y^* = I_h/s, x^* = I_w/s$  (and $M = (y_1,x_1)$) for a single pooling layer.
Then the dimensions of $W$ can be written as $W=(y_1,x_1,y^*,x^*,k)$.
As can be seen in Figure~\ref{funwrape} we have to bring the inner $x$ and $y$ dimensions to the right and the outer ones to the left.
The pooling is the inner dimension as it moves pixel by pixel. Thus, $O$ can be written as $O= (I_h = (y^*,y_1) ,I_w = (x^*,x_1),k) \overset{reshape}{=} (y^*,y_1,x^*,x_1,k)   $. 

For the more general case of $n$ pooling layers  $y^* = I_h/(s_1...s_n)$, $x^* = I_w/(s_1...s_n)$, \\  $W = (M=(y_n,x_n ... y_1,x_1),y^*,x^*,k)$ 
we have to get to  $O=(y^*,y_n,...,y_1,x^*,x_n,...,x_1,k)$.
An efficient and generic way to do this transformation only with transpose  and reshape operations is as follows:
first we have to transpose $(M,y^*,x^*,k)$ to $(y^*,x^*,M,k)$ . The naive way requires two transpose operations.
However, with reshaping it can be done in one: $(M,y^*,x^*,k) \rightarrow (M,f^*,k) \rightarrow (f^*,M,k) \rightarrow (y^*,x^*, M,k)$.
Note that reshaping requires barely any runtime as it is just a reinterpretation of the memory. Then, we do the following:
\begin{enumerate}\itemsep0.0pt
 \item  start:  $(y^*,x^*, M,k)$
 \item reshape $M$ to its contents : $(y^*,x^*, y_n,x_n ... y_1,x_1,k)$
 \item do $n$ times 
 \item ~~transpose: (dim 2, dim 3)
 \item ~~reshape: fuse (dim 1, dim 2), fuse (dim 3, dim 4)
 \item end
\end{enumerate}
\vspace{-0.2cm}
After performing this algorithm we have determined $O$.

\subsection{Use in practice at the example of torch}
In practice things can differ from theory. In Torch,\footnote{http://torch.ch/} for instance, the channel dimension is before the spatial 
dimension i.e. $W=(M,k,y^*,x^*)$.
Still we can do our initial step with only one transpose operation: $(M,k,y^*,x^*) \rightarrow (M,f^*) \rightarrow (f^*,M) \rightarrow (k,y^*,x^*,M)$.
To obtain a dimension for $M$ we use an unsqeeze layer before the first CNN layer. An example torch implementation can be found in the appendix.

\section{Experiments}\label{exp}
In this section we present benchmark results of our improved network architecture $C_I$ compared to $C_P$ running on all patches of an image.
The experiments are performed on a  GeForce GTX TITAN X. Readers who want to replicate our modify our experiments can use our benchmark code provided in the supplementary material. 
We do not count the time to extract patches from the image for $C_P$, but only the pure network processing time, although including this preprocessing step only
required for $C_P$ would increase our speedup even more. 
As can be seen in Table~\ref{f_subscales}, the execution time of $C_P$ roughly scales (as expected) linearly with the image pixels.
$C_I$ on the other hand barely takes more time for larger images. We think that this is due to overhead and because the GPU cores are not fully occupied by $C_I$.
In theory it should for large images also scale linearly with image pixels. 
On the other hand, the memory consumption of $C_I$ increases nearly linearly. 
If not enough memory is available the input image can be split into parts and each part can be processed individually.
This requires some -- in practice barley relevant -- computational redundancy. 
The memory requirement of $C_P$ depends on the number of patches processed in parallel.
Processing more patches is usually faster, but very large amounts do not have much influence anymore, as GPU cores are limited (still we used large amounts for the fastest possible runtime).

As can also be seen in the table, architectures like~\cite{bailer2016cnn} that perform heavy convolution can be speed up a little more than
architectures that perform heavy pooling like~\cite{simo2015discriminative, zagoruyko2015learning}.

\begin{table}[t]
\small
 \centering
 \begin{tabular}{|c|c|c|c|c|c|c|c|c|}
 \hline
 Architecture  & Image Size & $C_P$ & $C_I$ & Speedup & memory $C_I$\\
  \hline
Simo-Serra et al.~\cite{simo2015discriminative}  & 72 x 48 &   1.25 s &  0.099 s  & 12.6 times & 348 MB \\
  \hline
Simo-Serra et al.~\cite{simo2015discriminative}  & 360 x 240  & 28.42 s&  0.103 s & 275 times & 1142 MB\\
 \hline
Simo-Serra et al.~\cite{simo2015discriminative}  & 720 x 480   & 112.76 s &  0.116 s & 972 times & 3448 MB \\
  \hline
Simo-Serra et al.~\cite{simo2015discriminative}  & 1080 x 720   & 252.97 s&  0.117 s & 1550 times & 7222 MB \\
   \hline
Zagoruyko et al.\cite{zagoruyko2015learning}(Siamese $L_2$)& 1080 x 720    & 223.55 s& 0.127 s & 1760 times & 11495 MB \\
  \hline
Bailer et al.\cite{bailer2016cnn} (fast arch) & 864 x 576   & 363.45 s&  0.113 s & 3216 times & 9850 MB \\
    \hline
 \end{tabular}
 \vspace{0.1cm}
 \caption{Speed benchmark for $C_P$ and $C_I$. The latter performs much faster especially on larger images.
 We provide source code for the benchmarks in our supplementary material to prove reproducibility. For $C_P$ memory is not a big deal (see text).}
\label{f_subscales}
\end{table}

\section{Conclusion}
In this paper, we presented a novel approach to convert nearly arbitrary CNN architectures for fast execution on the whole image in a sliding window manner.
We showed that with our approach significant speedups can be achieved --
dense feature maps of state-of-the-art CNN based feature descriptors can be created in barely more than 0.1s for a whole image.
This is very interesting considering that CNN based features are nowadays more robust than most computationally intensive traditional features --
and with our approach, it is now also possible to compute them very fast densely.

By providing a straightforward implementation, we hope to convince authors of future works that there is no need to perform unnecessary,
redundant calculations or to avoid pooling layers.
This is particularly relevant for dense-feature computations on the whole image.

\appendix 
\section*{Appendix: Torch example}
Here we present some functions that allow to convert an arbitrary CNN $C_P$ to an architecture $C_I$.
An example implementation of~\cite{simo2015discriminative} shows the usage of these functions. 
Note:~\cite{simo2015discriminative} uses mean-pooling, but for simplicity we use the much more popular max-pooling in our example.
Using mean pooling instead is straightforward.

\begingroup

\small
\begin{lstlisting}
-------------- Torch functionality of our approach:------------

-- Adds padding to the image and adds the dimesion M
function multiPoolPrepare(patchY,patchX)
 padx = patchX-1
 pady = patchY-1
 local net = nn.Sequential()
 net:add(nn.Padding(2,-torch.ceil(pady/2)))
 net:add(nn.Padding(3,-torch.ceil(padx/2)))
 net:add(nn.Padding(2,torch.floor(pady/2)))
 net:add(nn.Padding(3,torch.floor(padx/2)))
 net:add(nn.Unsqueeze(1))
 return net
end

-- Layer prepares unwarping. Added after the network
function unwarpPrepare()
 local net = nn.Sequential()
 net:add(nn.View(-1):setNumInputDims(3)) --puts dim 2,3,4 to one dim
 net:add(nn.Transpose( {1,2} ))
 return net
end

-- The actual unwarping. See example for usage
-- curImg = imgSize /(all still existing strides)
function unwarpPool(outChans, curImgW, curImgH, dW, dH)
  local net = nn.Sequential()
  net:add(nn.View(outChans,curImgH,curImgW,dH,dW,-1))
  net:add(nn.Transpose({3,4})) -- {3,4} not {2,3} as k is first dim 
  return net
end

-- Replaces normal maxpooling
function multiMaxPooling(kW,kH,dW,dH)
  local c1 = nn.DepthConcat(1)
  for i = 0,dH-1 do
    for j = 0,dW-1 do
        c1:add(nn.SpatialMaxPooling(kW,kH,dW,dH,-j,-i))
    end
  end
  return c1
end


---------------------- Examaple usage:----------------------

sL1 = 2 -- stride 1. pooling layer
sL2 = 3 -- stride 2. pooling layer
sL3 = 4 -- stride 3. pooling layer

-- image height and width should be multiples of sL1*sL2*sL3
-- if this is not the case right/downwards padding should be added.
imH = ... -- image height 
imW = ... -- image width  

pH = 64 -- patch height
pW = 64 -- patch width

outChans = 128 -- output channels (last layer)

-- The patch network (C_P):
net1 =  nn.Sequential()
net1:add(nn.SpatialConvolution(3, 32, 7, 7))
net1:add(nn.SpatialMaxPooling(sL1 ,sL1 ,sL1 ,sL1 ))
net1:add(nn.TanH())
net1:add(nn.SpatialConvolution(32, 64, 6, 6))
net1:add(nn.SpatialMaxPooling(sL2 ,sL2 ,sL2 ,sL2 ))
net1:add(nn.TanH())
net1:add(nn.SpatialConvolution(64, outChans, 5, 5))
net1:add(nn.SpatialMaxPooling(sL3 ,sL3 ,sL3 ,sL3 ))
net1:add(nn.TanH())


-- The image network (C_I):
net2 =  nn.Sequential()
net2:add(multiPoolPrepare(pH,pW))  
net2:add(net1.modules[1])
net2:add(multiMaxPooling(sL1,sL1,sL1,sL1))
net2:add(net1.modules[3])
net2:add(net1.modules[4])
net2:add(multiMaxPooling(sL2,sL2,sL2,sL2))
net2:add(net1.modules[6])
net2:add(net1.modules[7])
net2:add(multiMaxPooling(sL3,sL3,sL3,sL3))

net2:add(unwarpPrepare())
net2:add(unwarpPool(outChans, imH/(sL1*sL2*sL3), imW/(sL1*sL2*sL3),
sL3, sL3))
net2:add(unwarpPool(outChans, imH/(sL1*sL2), imW/(sL1*sL2),sL2,sL2))
net2:add(unwarpPool(outChans, imH/sL1, imW/sL1 , sL1, sL1) )
net2:add(nn.View(-1,imH,imW) 

\end{lstlisting}
\endgroup

\bibliography{egbib}

\begin{thebibliography}{13}
\providecommand{\natexlab}[1]{#1}
\providecommand{\url}[1]{\texttt{#1}}
\expandafter\ifx\csname urlstyle\endcsname\relax
  \providecommand{\doi}[1]{doi: #1}\else
  \providecommand{\doi}{doi: \begingroup \urlstyle{rm}\Url}\fi

\bibitem[Bailer et~al.(2016)Bailer, Varanasi, and Stricker]{bailer2016cnn}
Christian Bailer, Kiran Varanasi, and Didier Stricker.
\newblock Cnn-based patch matching for optical flow with thresholded hinge
  loss.
\newblock \emph{arXiv preprint arXiv:1607.08064}, 2016.

\bibitem[Barnes et~al.(2010)Barnes, Shechtman, Goldman, and
  Finkelstein]{barnes2010generalized}
Connelly Barnes, Eli Shechtman, Dan~B Goldman, and Adam Finkelstein.
\newblock The generalized patchmatch correspondence algorithm.
\newblock In \emph{ECCV}, pages 29--43. Springer, 2010.

\bibitem[Bay et~al.(2006)Bay, Tuytelaars, and Van~Gool]{bay2006surf}
Herbert Bay, Tinne Tuytelaars, and Luc Van~Gool.
\newblock Surf: Speeded up robust features.
\newblock \emph{Computer vision--ECCV 2006}, pages 404--417, 2006.

\bibitem[Bromley et~al.(1993)Bromley, Bentz, Bottou, Guyon, LeCun, Moore,
  S{\"a}ckinger, and Shah]{bromley1993signature}
Jane Bromley, James~W Bentz, L{\'e}on Bottou, Isabelle Guyon, Yann LeCun, Cliff
  Moore, Eduard S{\"a}ckinger, and Roopak Shah.
\newblock Signature verification using a siamese time delay neural network.
\newblock \emph{International Journal of Pattern Recognition and Artificial
  Intelligence}, 7\penalty0 (04):\penalty0 669--688, 1993.

\bibitem[Gadot and Wolf(2016)]{gadot2016patchbatch}
David Gadot and Lior Wolf.
\newblock Patchbatch: a batch augmented loss for optical flow.
\newblock In \emph{Proceedings of the IEEE Conference on Computer Vision and
  Pattern Recognition}, pages 4236--4245, 2016.

\bibitem[Ge et~al.(2015)Ge, McCool, Sanderson, and Corke]{ge2015modelling}
ZongYuan Ge, Chris McCool, Conrad Sanderson, and Peter Corke.
\newblock Modelling local deep convolutional neural network features to improve
  fine-grained image classification.
\newblock In \emph{Image Processing (ICIP), 2015 IEEE International Conference
  on}, pages 4112--4116. IEEE, 2015.

\bibitem[Girshick(2015)]{girshick2015fast}
Ross Girshick.
\newblock Fast r-cnn.
\newblock In \emph{Proceedings of the IEEE International Conference on Computer
  Vision}, pages 1440--1448, 2015.

\bibitem[Han et~al.(2015)Han, Leung, Jia, Sukthankar, and
  Berg]{han2015matchnet}
Xufeng Han, Thomas Leung, Yangqing Jia, Rahul Sukthankar, and Alexander~C Berg.
\newblock Matchnet: unifying feature and metric learning for patch-based
  matching.
\newblock In \emph{Computer Vision and Pattern Recognition (CVPR)}, 2015.

\bibitem[Lowe(1999)]{lowe1999object}
David~G Lowe.
\newblock Object recognition from local scale-invariant features.
\newblock In \emph{International Conference on Computer Vision (ICCV)}, 1999.

\bibitem[Simo-Serra et~al.(2015)Simo-Serra, Trulls, Ferraz, Kokkinos, Fua, and
  Moreno-Noguer]{simo2015discriminative}
Edgar Simo-Serra, Eduard Trulls, Luis Ferraz, Iasonas Kokkinos, Pascal Fua, and
  Francesc Moreno-Noguer.
\newblock Discriminative learning of deep convolutional feature point
  descriptors.
\newblock In \emph{International Conference on Computer Vision (ICCV)}, 2015.

\bibitem[Simonyan et~al.(2014)Simonyan, Vedaldi, and
  Zisserman]{simonyan2014learning}
Karen Simonyan, Andrea Vedaldi, and Andrew Zisserman.
\newblock Learning local feature descriptors using convex optimisation.
\newblock \emph{Pattern Analysis and Machine Intelligence (PAMI)}, 36\penalty0
  (8):\penalty0 1573--1585, 2014.

\bibitem[Zagoruyko and Komodakis(2015)]{zagoruyko2015learning}
Sergey Zagoruyko and Nikos Komodakis.
\newblock Learning to compare image patches via convolutional neural networks.
\newblock In \emph{Computer Vision and Pattern Recognition (CVPR)}, 2015.

\bibitem[Zbontar and LeCun(2016)]{zbontar2016stereo}
Jure Zbontar and Yann LeCun.
\newblock Stereo matching by training a convolutional neural network to compare
  image patches.
\newblock \emph{Journal of Machine Learning Research}, 17\penalty0
  (1-32):\penalty0 2, 2016.

\end{thebibliography}
\end{document}